\documentclass{article}

\usepackage{arxiv}

\usepackage[utf8]{inputenc} 
\usepackage[T1]{fontenc}    
\usepackage{hyperref}       
\usepackage{url}            
\usepackage{booktabs}       
\usepackage{amsfonts}       
\usepackage{nicefrac}       
\usepackage{microtype}      
\usepackage{lipsum}
\usepackage{graphicx}
\usepackage{amssymb}

\usepackage{amsmath}
\usepackage{inputenc}

\usepackage{subcaption}
\usepackage{rotating}
\usepackage{tikz}
\usepackage{lscape}
\usetikzlibrary{positioning}
\usepackage{array,calc}
\newlength\lena \newlength\lenb \newlength\lenc \newlength\lend
\newcolumntype{P}[1]{>{\centering\arraybackslash}p{#1}} 

\title{A Differentially Private Multi-Output Deep Generative Networks Approach for Activity Diary Synthesis}

\author{
  Godwin Badu-Marfo \\
  Laboratory of Innovations in Transportation\\
 Ryerson University\\
 Toronto, Canada \\
 \texttt{gbmarfo@ryerson.ca} \\
   \And
  Bilal Farooq \\
  Laboratory of Innovations in Transportation\\
  Ryerson University\\
  Toronto, Canada \\
  \texttt{bilal.farooq@ryerson.ca} \\
   \And
  Zachary Patterson \\
  Department of Geography, Planning and Environment\\
  Concordia University\\
  Montr\'eal, Canada \\
  \texttt{zachary.patterson@concordia.ca} \\
}

\begin{document}
\maketitle

\begin{abstract}

In this work, we develop a privacy-by-design generative model for synthesizing the activity diary of the travel population using state-of-art deep learning approaches. This proposed approach extends literature on population synthesis by contributing novel deep learning to the development and application of synthetic travel data while guaranteeing privacy protection for members of the sample population on which the synthetic populations are based. First, we show a complete de-generalization of activity diaries to simulate the socioeconomic features and longitudinal sequences of geographically and temporally explicit activities. Second, we introduce a differential privacy approach to control the level of resolution disclosing the uniqueness of survey participants. Finally, we experiment using the Generative Adversarial Networks (GANs). We evaluate the statistical distributions, pairwise correlations and measure the level of privacy guaranteed on simulated datasets for varying noise. The results of the model show successes in simulating activity diaries composed of multiple outputs including structured socio-economic features and sequential tour activities in a differentially private manner.

\end{abstract}


\section{Introduction}
\label{intro}

Activity-based travel demand models have become staple in the academic transportation modeling literature, and increasingly in transportation decision-making in recent times. Using these models, transportation planners and stakeholders study the behaviour of agents (i.e. households and individuals) that influences their choices of daily activity participation and travel. Activity-based models require spatially and temporally granular representations of a person's trip activities including where and when activities take place, and how (which travel mode) they get to activity locations. These microsimulation models necessarily rely heavily on disaggregate, individual-level information (i.e. microdata). In practice, it is difficult to obtain disaggregate travel information on agents because of the high cost of data collection for large populations, as well as the potentially privacy compromising effects of inadvertent releases or intentional publishing of such data. 

As a solution to address the lack of accessibility to, and completeness of, microdata, population synthesis is used to generate synthesized representations of true populations based on samples of disaggregate data that are characterized by the same joint distribution of variables of the real population \cite{muller2010population, farooq2013simulation}. This technique is appealing for generating data surrogates with properties that conform to the underlying distribution of the population but which only require data from a small sample from the population of interest. Popular methods of population synthesis including re-weighting and matrix fitting do not produce agent-based samples but rather samples of prototypical weighted individuals, and hence a post-simulation is required in cases where population synthesis is linked to agent-based samples where individuals are drawn from the weighted samples \cite{muller2010population, zhang2019connected}. An increasingly popular approach to population synthesis, simulation, solves some of the drawbacks of the re-weighting and matrix fitting models. Simulation-based methods have proved effective for high-dimensional synthetic generation and provide a systematic way for imputing or interpolating data \cite{saadi2016hidden, farooq2013simulation}. Farooq et al. 2013 \cite{farooq2013simulation} used this approach to generate a synthetic micro population for Brussels in Belgium where complete data for the population was not available. All of these methods of population synthesis suffer the drawback of scalability due to the ``curse of dimensionality'' and computational complexity \cite{farooq2013simulation, borysov2019scalable}.

These traditional population synthesis approaches have shown success in recreating weighted samples from aggregate census data, but have not been used to generate representations of complete travel diaries due to issues around computational complexity and scalability. Outside of transportation, Generative Adversarial Networks (GANs) have proved capable of estimating complex joint distributions,  hitherto intractable for large training sets and complex data types like video, images, sound, etc. In the GANs framework, a generative model is set against an adversary, or a ``discriminative'' model that learns to distinguish fake observations produced by the generative model from real data observations. In the transportation literature, Borysov et al. \cite{borysov2019scalable} demonstrated the use of GANs for simulating the socio-demographic characteristics of synthetic agents for a travel model. Unlike here, their work only synthesized tabular data and did not incorporate the sequential data generation. The capability of GANs to reproduce faithful representations of a population, however, could lead to information leakage \cite{shokri2017membership, hayes2019logan} on training data points that threatens the privacy of respondents. The potential for compromising the privacy of synthesis seed sample members raises research interest in developing newer synthesis approaches that exhibit privacy-by-design capabilities. To this end, recent work \cite{abadi2016deep, shokri2015privacy, chamikara2019local} has focused on developing deep learning approaches that protect sensitive information by training in a differentially private manner such that the privacy of synthesis seed sample members is not compromised. Abadi et al. \cite{abadi2016deep} have demonstrated the possibility of training a model in a differentially private way that relies on the Differentially Private Stochastic Gradient Descent (DP-SGD). This privacy-sensitive training can control the confidence with which an adversary can learn or infer information about an individual from a sample, and can indeed control this through a parameter, epsilon ($\epsilon$) that defines the level of privacy guaranteed. 

In this paper, we leverage the potential of previous work to solve three problems: First, we want to synthesize a complete activity diary  (based on socioeconomic attributes and a snapshot of longitudinal activity sequences of a sample) to synthesize travel diaries for a synthetic population. To the best of our knowledge, most of the previous synthesis approaches only focused on synthesizing tabular data i.e. socioeconomic attributes. Second, we explore training the generative model in a differentially private manner as a step to protect sensitive information of individuals in the underlying training data. Finally, build and deploy a novel generative mechanism that adopts state-of-art deep learning techniques like Generative Adversarial Networks. The population synthesized using our proposed methodology can be directly used as the base-input for the modern activity-based travel demand modelling systems.
   
As such, the key contributions of our work include:

\begin{enumerate}
    \item We present a novel generative model that is capable of estimating the joint distribution of socioeconomic variables of travel agents and simultaneously learn the agent activity sequences from an Origin-Destination (OD) survey, while incorporating parameters to guarantee privacy of information leakage about a person who participated in a survey.
    \item We adopt a privacy-by-design generative approach to limit the influence of gradients learnt on training points resulting in privacy preservation for the simulated population.
    \item We experiment with a differentially private training of the generative model with varying degrees of noise to evaluate the effect on the statistical distribution of the synthesized representations. 
    \item To the best of our knowledge, this is the first work using Generative Adversarial Networks to synthesize a complete activity diary of agents having multiple outputs of socioeconomic characteristics and a complete activity chain in a single model.
\end{enumerate}

The rest of the paper is organized as follows: the next section presents review of the relevant literature, followed by a section that describes the framework architecture of the generative model. A methodology section describes the data processing steps and we then define the evaluation metrics before presenting an analysis of results. We finish the paper by explaining our conclusions and future directions for the research.

\section{Literature Review}

Population synthesis approaches have been common in trip-based modelling over the years to estimate synthetic members of a population in cases where data on travel agents (i.e. individuals and households) are not available. Using as inputs census aggregates and microdata samples of individuals in a study region, synthetic people and households can be simulated that possess similar travel characteristics of the true population. Synthesis approaches are broadly classified into three categories; re-weighting, matrix fitting, and simulation based approaches \cite{tanton2014review}. The re-weighting methods adjust weight factors of surveys to create samples that represent subregions rather than the entire summation of the population aggregates, in effect, applying non-linear optimization to estimate weights \cite{bar2009estimating, daly1998prototypical, harland2012creating}. The methods of matrix fitting involves de estimation of expansion factors that are the ratio between a starting solution and the final matrix. The Iterative Proportion Fitting (IPF) proposed by Deming and Stephan \cite{deming1940least} and the Maximum Cross-Entropy \cite{guo2007population} are known implementations of the matrix fitting method and are referred to as \textit{deterministic models}. These deterministic models do not produce agent-based samples but rather a sample of prototypically weighted agents \cite{borysov2018scalable}. Duguay et al. \cite{duguay1976synsam} first introduced the IPF method to synthesize household survey data in the transportation literature. Similarly, Beckman et al. \cite{beckman1996creating} created a synthetic population for use with  TRANSIMS \cite{laron1996transims} using census cross tabulations and micro samples. Adopting fitting methods for large dimensional data becomes computationally and memory intensive. Simulation-based methods solve some of the drawbacks of the deterministic models and are capable of estimating the joint distribution of population data with a full set of attributes from which new members can be recreated through sampling. Sun and Erath \cite{sun2015bayesian} proposed a Bayesian Network approach, a popular implementation of the simulation-based approach. Similarly, Sun et al.\cite{sun2006bayesian} showed that a Bayesian network approach is effective for traffic flow modelling and forecasting while performing experiments on urban vehicular traffic flow data for Beijing. However learning of the graph structure of bayesian networks for large datasets can be computationally intensive \cite{borysov2019scalable}.

Deep generative models have evolved lately to reproduce realistic and near-true synthetic representations that perform effectively in dealing with the complex computations required for synthesizing agents. The most popular variants of deep generative models are the Generative Adversarial Networks (GANs)~\cite{goodfellow2014generative} and Variational Autoencoders (VAE) \cite{kingma2013auto}. Similar to the simulation-based approaches, these deep generative models are capable of estimating the joint distribution of the underlying data where new samples can be generated. Choi et al.~\cite{choi2017generating} proposed a generative model that combines auto-encoders with GANs to synthesize private electronic health records in generating binary and count variables in health datasets. Park et al. \cite{park2018data} proposed a \textit{table-GAN} to synthesize tabular data using a hinge-loss privacy control mechanism. In their approach, they showed a compatible model for anonymization where sensitive attributes are maintained without change. Neural sequence generation has been well studied since the advent of Recurrent Neural Networks (RNN) \cite{pascanu2013construct} and Long Short Term Memory (LSTM)~\cite{hochreiter1997long}. RNNs have shown incredible results in capturing long-term dependencies but as Bengio \cite{bengio2015scheduled} discussed, fitting the distribution of observed data does not mean generating satisfactory text because of ``exposure bias" \cite{lu2018neural}. Solutions proposed to address these limitations include the concept of reinforcement learning and GANs to generate acceptable sequences. SeqGAN \cite{yu2017seqgan} was proposed as a language model for the generation of sequences using Reinforcement Learning (RL). As an RL, the authors assume the state is defined as the tokens generated, and its action being the next token to be generated. Due to the discrete outputs of text sequences and the requirement to evaluate partial sequences generated, the authors assume the generator of the network as a stochastic parameterized policy. Using stochastic policy, the REINFORCE \cite{ranzato2015sequence} algorithm, allows different actions to be sampled during training and to derive a robust estimate of the policy. Both the generator and discriminator are pretrained on real and fake data prior training with policy gradients. During training they implement Monte Carlo rollouts in order to get a useful loss signal for each word. Subsequent work demonstrated text generation without pretraining with RNNs \cite{press2017language}. 

While GANs and other generative models have been successful for reproducing identical copies of the true population, there is a risk of information leakage to an adversary who could infer if a person partook in the training dataset. Such concerns have motivated recent research work into developing privacy-by-design techniques such as differentially private training in deep learning \cite{abadi2016deep}. Abadi et al. studied a gradient clipping method that imposed privacy during the training of the neural network. Shokri and Shmatikov \cite{shokri2015privacy} proposed a multi-party privacy preserving neural network with a parallelized and asynchronous training procedure. In the work of Phan et al.\cite{phan2017preserving}, the authors developed private convolutional deep belief networks (CBDNs) by leveraging the functional mechanism to perturb the energy-based functions of conventional CBDNs.   

In the next sections, we provide a brief definition of topics including necessary to understand our proposed modeling approach. The topics include: deep generative modelling and differential privacy.

\subsection{Deep Generative Modelling}
Deep generative models have evolved out of artificial neural networks \cite{hinton1992neural} where they have been used successfully to reproduce realistic images and translations, while exhibiting outstanding performance and computational effectiveness. Notable deep generative models are the Variational AutoEncoder (VAE) \cite{kingma2013auto} and Generative Adversarial Networks (GANs) \cite{goodfellow2014generative}. Both generative models have shown promising results in estimating the joint distribution of underlying data, a property that is important for simulation-based population synthesis techniques like Bayesian Networks.

\begin{figure}[!h]
    \centering
    \includegraphics[width=\textwidth]{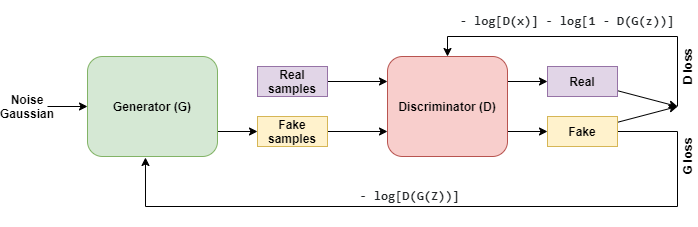}
    \caption{The architecture of GANs}
    \label{fig:gan}
\end{figure}

Ian Goodfellow \cite{goodfellow2014generative} proposed the Generative Adversarial Network that simulates a two player game composed of ``Generator'' and ``Discriminator'' networks. Generator networks learn to synthesize samples from latent space that are intended to mimic real sample data. The Discriminator network is programmed to distinguish between synthesized and real sample data, with updated weights being sent back to the generator. Models for both networks are implemented as multi layer perceptrons \cite{hinton2007learning}. During model training, the Discriminator gets better at distinguishing real samples from fake synthesized samples, while the generator improves on generating samples that are close to the real samples until a Nash equilibrium is achieved where each model reaches its peak ability to thwart the other's goal. 

The objective function of GANs is defined by:

\begin{flushleft}

\textbf{Definition 1 (Objective function): }

The \textit{objective function} of the Generative Adversarial Networks \cite{goodfellow2014generative} is:
\begin{equation} \label{eq1}
\underset{G}{min} \ \underset{D}{max} \ V(D,G)= \mathbb{E}{_{x \sim p_{data}(x)}}[log \ D(x)] + \mathbb{E}{_{z \sim p_{z}(z)}}[log(1 - D(G(z)))] 
\end{equation}

\end{flushleft}

where \textit{V (G, D)} is the value function of a two-player minimax game, \textit{D(x)} represents the probability that x came from the true data, p$_g$ is the generator’s distribution over the real data \textit{x}, p$_z$(z) is a prior on noise variable (z) that maps to data space as G(z), where G is a differentiable function represented by a multilayer perceptron.

Equation \ref{eq1} derives the objective function that implies that if the input of the Discriminator is sampled from the true distribution, then maximize the output of D(x) to 1 whereas if the input comes from the Generator then D(G(z)) should minimize the output of the objective function. In this regard, the network seeks to maximize the parameters of the Discriminator using Gradient Ascent while minimizing the parameters of the Generator using Gradient Descent. The training process halts when a Nash equilibrium is reached so that the Discriminator is unable to distinguish true or fake samples.

In the work of Choi et al. \cite{choi2017generating}, the authors proposed a model that combines auto-encoders with GANs to synthesize private electronic health records. The results of their model simulated binary and count variables in the context of health datasets. Similar work by Park et al. \cite{park2018data} proposed a \textit{table-GAN} to simulate tabular data using a hinge-loss privacy control mechanism that is suitable for anonymization of sensitive attributes. Borysov et al. \cite{borysov2019scalable} has shown a simulation of micro-agents from a large Danish activity diary to estimate the joint distribution of the underlying data using the VAE model. In our approach, the GANs architecture will be optimized for high performance throughput making it capable of learning all training data records in order to avoid the challenge of sampling zeros, referring to agents that are omitted from the training samples but exist in the real population.

\subsection{Differential Privacy}
The concept of differential privacy \cite{dwork2011differential} assures nothing new can be learnt on the statistical output of a query mechanism given that a record of information on an individual is added or removed from two neighboring databases. In this sense, a privacy guarantee is provided in the query function, which is not the case for other anonymization techniques like k-Anonymity \cite{sweeney2002k}. Differential privacy limits a constraint on the processing of data such that the output of executing a query mechanism on two adjacent databases is approximately similar. The functional mechanisms of achieving differential privacy include the approach of adding Laplacian noise \cite{dwork2008differential}, the exponential mechanism \cite{mcsherry2007mechanism}, and the functional perturbation approach \cite{chaudhuri2009privacy}. According to Dwork et al. \cite{dwork2008differential}, a randomized algorithm \textit{M} fulfills $\epsilon$-differential privacy if, for any adjacent databases \textit{d} and \textit{d'} differing at most one element, and for any output O of \textit{M} is formally defined by:

\begin{flushleft}

\textbf{Definition 2 ($\epsilon$-Differential Privacy): }

The formal definition of \textit{$\epsilon$-Differential Privacy} is given by:
\begin{equation} \label{eq2}
Pr(M(d)=O)\leqslant e^{\epsilon}Pr(A(d^{\prime})=O)
\end{equation}

\end{flushleft}

The privacy budget defined by parameter epsilon($\epsilon$) defines the difference between adjacent databases \textit{d} and \textit{d$^\prime$}, differing by only one observation. A controlled random noise is sampled from a Laplace distribution that is added to the query output of the function mechanism to achieve differential privacy.

\subsection{Deep learning with differential privacy}
As a step towards implementing differentially private training, we adopt the approach by Abadi et al. \cite{abadi2016deep} in our work. Abadi et al. developed a technique to train deep learning models in a differentially private manner. In their approach, random noise is sampled from a Gaussian distribution and added to the gradients of parameters of the neural network. The addition of noise to the computed gradients limits the influence that any particular input data can have on the final model. The steps for differential privacy training are as follows:

 \begin{itemize}
    \item Sample a minibatch of training data (x, y) where x is the input and y is the label. 
    \item Compute loss L($\theta$, x, y) defined as the difference between the model's prediction $\theta(x)$ and label y where $\theta$ represents the parameters of the model.
    \item Compute the gradient of the loss L($\theta$, x, y) with respect to the parameters $\theta$.
    \item For each training example, clip gradients in the minibatch to an upper bound defined by the maximum euclidean norm. 
    \item Add random noise sampled from a Gaussian distribution to the clipped gradients and update parameters.
\end{itemize}
 
The model training with differential privacy gives a sanitized model gradient in which the influence of input data is bounded thus achieving privacy. The bounded gradients are used to train the model while updating the weights. 

\subsection{Membership Inference Attacks Against Generative Models}

The approach of differentially-private training promises of privacy protection for samples used as training data points. In order to evaluate the models robustness and susceptibility to adversarial attacks, Membership Inference Attacks (MIA) are implemented. The notion of Membership Inference Attack (MIA) was proposed by Shokri \cite{shokri2017membership}, as a privacy attack mechanism to measure the robustness of machine learning privacy protecting algorithms against adversarial attacks. (Attacks that attempt to infer the identity or uniqueness of individuals from anonymized data.) An attack evaluates the prediction score of a model when the input data point is sampled from the training set rather than the validation set. The MIA comes in two forms: \textit{black-box} and \textit{white-box} attacks. Black-box attacks assume an adversary can only make queries to the target model under attack but has no access to the internal parameters of the model \cite{shokri2017membership}. On the other hand, white-box attacks assume the adversary has the parameters of the trained model and can make queries to the target model. We adopt a white-box attack approach in this work because it is simple to implement and efficient. In a GAN setting, the adversary is only given access to the discriminator of the trained GAN model and considers a setting where the model parameters are leaked following a data breach. The trained model determines if a record was part of the training set, consequently the attack analyzes the danger in identifying with high confidence if an observation was used in the training. Additionally, the adversary is assumed to have knowledge of the proportion of the dataset that is used for training, but no other subsequent information is known about the training set. This attack is implemented by deriving the predictive scores of the discriminator of the target model given a sample of the training set.  In a non-private trained model, the output of the attack should score lower probabilities (i.e., close to 0) for validation sets and high probabilities (close to 1) for training sets. On the other hand, private trained models should not output scores that distinguish training sets from validation sets.

As a sequel to earlier discussion, population synthesis approaches suffer the drawback of scalability to high dimensional data that are computationally expensive. To the best of the authors' knowledge, this is the first work to simulate both the tabular and sequential attributes of activity diary. The methodology of this work seeks to address these limitations using novel deep learning methods. In the next section, we discuss the methodology.

\section{Methodology}
In this section, we first introduce the problem definition to establish the goal of the research. We continue in the subsequent subsections to give a detailed description of the proposed architecture for synthesizing tabular socioeconomic variables and longitudinal activity sequences of location. 

\subsection{Problem definition}
First, we assume, \textit{X} to be the training data containing sensitive travel information on individuals. The training data is comprised of structured socio-economic variables characterized by a set of basic attributes $X=(x_1, x_2, x_3, x_4, ... x_n)$ where n is the number of variables, and a longitudinal sequence of time-ordered trip activities including trip purpose, departure time, and geographic coordinates of origins and destinations.

A generative model, \textit{M} is trained on the private data and new data, \textit{X$^\prime$}, is sampled from the model. In practice, the true data distribution of the population, $p_{data}(X)$ is unknown but it is approximated here empirically using a sample. For the purpose of data synthesis, we use GANs as a framework to estimate the $p_{data}(X)$ and subsequently draw samples from it. In order to maintain privacy protection for participants in a travel survey, the generative model will be required to prevent an adversary from recovering with a high degree of confidence that an individual participated in the training data of the generative model, or prevent the adversary from inferring sensitive information about an individual based on the output of the model. In this sense, the goal of the proposed differentially private generative model is to synthesize a complete activity diary with high utility while guaranteeing privacy protection on training data.

\subsection{Differentially Private Composite Travel Generative Adversarial Network}

The proposed Differentially Private Composite Travel Generation Adversarial Network (DP-CTGAN) is a novel generative model that is designed to accept input from multiple data types (i.e. tabular and sequences) and is capable of estimating the joint distribution of data inputs through a shared hidden layer, and subsequently generate new private samples from the generative model trained in a differentially private manner. The DP-CTGAN is composed of two neural networks; the Generator network, \textit{G} and discriminator network, \textit{D}. 

\begin{figure}[ht]
    \centering
    \includegraphics[width=\textwidth]{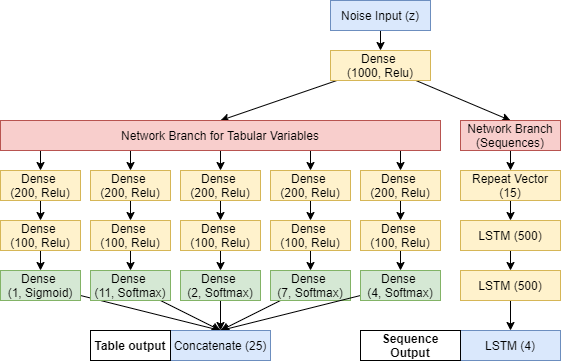}
    \caption{The Generator network of DP-CTGAN}
    \label{fig:generator}
\end{figure}

The Generator, \textit{G}, accepts as input a random noise that is sampled from a Gaussian distribution and which is fed into two branches of neural hidden layers. The first branch of the generator (\textit{G$_1$}), purposed for structured data learning is  made up of a series of multi-layer perceptrons (MLP) for each training variable that connects neurons for each layer to the neurons of the next layer. Each hidden layer is activated by a Rectified Linear Unit (ReLU) \cite{agarap2018deep} function which sets a lower bound of zero for negative inputs but returns same output for positive inputs. We apply a Sigmoid activation \cite{zhang2012weights} to the output layer for numeric variables (i.e. age), and a Softmax activation \cite{rahman2016towards} to the last hidden layer for categorical variables.

\begin{figure}[tbh]
    \centering
    \includegraphics[width=0.5\textwidth]{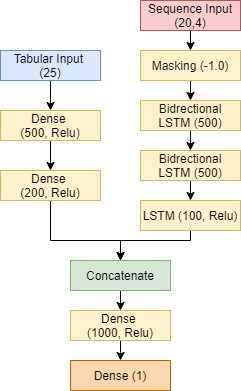}
    \caption{The Discriminator network of DP-CTGAN}
    \label{fig:discriminator}
\end{figure}

The second branch of the generator, \textit{G$_2$} is designed for sequential data learning. The first layer of \textit{G$_2$} is made up of a Keras \cite{gulli2017deep} Repeat Vector layer to repeat the incoming inputs in order to get hidden features for 20 future time-steps, the maximum length of each sequence. The output of the Repeat-Vector is fed to an LSTM layer with node size of 500 to extract features of previous time-steps. Two subsequent LSTM layers with node size of 500 are applied to the output of the time-steps such that the probabilities are well learnt for future sequence predictions. The last layer of the this branch is an LSTM with a size of 5, which is the number of the features for each time-step. The outputs of both branches \textit{G$_1$} and \textit{G$_2$} are merged into a shared output as the final output of the Generative model. 

The discriminator \textit{D} accepts inputs of the real tabular and sequential data as matrix of vector inputs. Similarly, \textit{D} is also made up of two branches. The first branch of the network, \textit{D$_T$} is an MLP made of two hidden layers with node sizes of 500 and 200 respectively. This branch accepts the input of the real tabular data and is purposed to learn the joint distribution of its input. The goal of the second branch, \textit{D$_S$} is to estimate the distributions of sequences by learning its weights. It is made up of two Bidirectional LSTM models with node sizes of 500 to learn the probabilities of sequences in both directions. The outputs are fed into an LSTM with node of 100. The outputs of both branches are then merged into a shared layer. The shared output is consequently fed into a Dense layer with node size of 1. 

In the design of the GAN, the generator does not have access to the real data but accepts an input of Gaussian noise. This makes it difficult to implement differential private training in the Generator. Contrarily, the Discriminator accepts the real data as its inputs hence making it suitable for training in a differentially private manner. To achieve privacy, we use the differentially private Stochastic Gradient proposed by \cite{abadi2016deep} to train the discriminator network as suggested by Xie et. al \cite{xie2018differentially} and the RMSProp Optimizer to train the Generator network. First, we introduce a clipping parameter to act as an upper bound on the L2-norm of each gradient update observed through the training. We also introduce a noise multiplier to control the ratio between the clipping parameter and the standard deviation of noise that is applied to each gradient update of the discriminator after clipping. In this paper, we use a range of noise multipliers from 0 to 4 as shown in table \ref{tab:noise}. The differentially private discriminator propagates its parameters to train a standard generator whose computed weights also become differentially private. In effect, newer samples predicted out of the generator, are guaranteed to be differential private.  

\begin{table}[!h]
\centering
\small
\begin{tabular}{|l|c|l|}
\hline
\textbf{Model} & \multicolumn{1}{l|}{\textbf{Noise Multiplier}} & \textbf{Description}             \\ \hline
WGAN (0.0)     & 0                                              & Model with no privacy noise      \\ \hline
WGAN (1.0)     & 1                                              & Model with noise multiplier of 1 \\ \hline
WGAN (2.0)     & 2                                              & Model with noise multiplier of 2 \\ \hline
WGAN (3.0)     & 3                                              & Model with noise multiplier of 3 \\ \hline
\end{tabular}
\caption{Noise multiples for differential private training}
\label{tab:noise}
\end{table}

\subsection{Case Study}
In this work, we evaluated DP-CTGAN on data from the 2013 Montr\'eal Origin-Destination (OD) Survey \cite{amt2013enquete}. The training data contained the activity diary of 10,000 individuals that were sampled out of the OD survey. The data included individual and household socio-economic variables such as gender, age, economic status, etc., and trip activity details such as activity location, time of departure, trip mode and purpose of travel. A list of the data available from the Montr\'eal OD Survey is shown in Table \ref{tab:survey-table}.

\begin{table}[!h]
\centering
\small
\begin{tabular}{lll}
\hline
\textbf{Column} & \textbf{Type} & \textbf{Description}                                \\ \hline
P\_AGE          & numeric       & Age of the respondent                               \\
P\_SEXE         & binary        & Gender of the respondent                            \\
P\_MOBIL        & categorical   & Whether the respondent is mobile                                  \\
P\_STATUT       & categorical   & Occupation status of respondent                     \\
PERMIT      & categorical   & Driving permit type of respondent                     \\
M\_DOMXCOOR     & geospatial    & X coordinate of residence      \\
M\_DOMYCOOR     & geospatial    & Y coordinate of residence      \\
D\_ORIXCOOR     & geospatial    & X coordinate of trip origins \\
D\_ORIYCOOR     & geospatial    & Y coordinate of trip origins\\
D\_DESXCOOR     & geospatial    & X coordinate of destinations\\
D\_DESYCOOR     & geospatial    & Y coordinate of destinations\\
D\_MOTIF     & categorical    & Trip purpose\\
\hline
\end{tabular}
\caption{Description of variables to be synthesized from 2013 Montr\'eal OD Survey}
\label{tab:survey-table}
\end{table}

\begin{table}[]
 \centering
\small
\begin{tabular}{|l|l|l|l|}

\hline
\textbf{Mean} & \textbf{Standard  deviation} & \textbf{Min} & \textbf{Max} \\ \hline
43            & 20                           & 5            & 95           \\ \hline
\end{tabular}
\caption{Summary statistics for numeric variable ``Age''}
\label{tab:num-summary}
\end{table}

\subsection{Data Preprocessing}
The OD survey data is made up of tuples of household and individual socio-economic variables as well as sequences of individual trips denoted by the coordinates of trip origins and destinations. The socio-economic variables have a fixed number of features for each individual comprising numerical (i.e. age), binary (sex) and categorical variables. On the other hand, trips of individuals have varying lengths of sequences having a minimum of three (3) locations and maximum of fifteen (15) location points. In this paper, we focus on the generation of home based trips, typically made up of a minimum of 3 locations (i.e., Origin-Destination-Origin).  As a first step towards training in neural networks, all variables are converted into normalized numeric representations that is recommended for achieving efficient training with neural networks. Binary and categorical variables are first encoded to integer indices and one-hot encoded \cite{potdar2017comparative}. Similarly, numeric variables (i.e. age, geographic coordinates) are scaled and normalized within a range from negative one (-1) and positive (+1). The summary statistics of variable ``age'' is shown in Table \ref{tab:num-summary}, which reports a minimum of 5 years and maximum of 95 years for respondents that partook in the survey.  While the objective of generative modelling is to recreate a synthetic copy of the true data, the encoding technique should be capable of being reversed or decoded to the initial state. In this work, we used Scikit-Learn algorithms \cite{bisong2019building}, label encoding and OneHot encoding which have reserve encoding capabilities.    

\subsection{Evaluation metrics and results}
In this section we empirically evaluate the performance of the generated synthetic representations of the population and their travel characteristics. We vary different noise levels of privacy to assess the privacy performance of the synthesized travel data. The evaluation is done using the following benchmarks.

\subsection{Similarity in statistical distribution}
Using this benchmark, we compare the statistical properties of the generated output to the training set to verify that their distributions are similar. A generated output should be appropriate for microsimulation estimations if aggregate queries on distributions are identical to the true distribution. To achieve this, we first sample from the marginal distribution of each variable $\pi(x_i)$ independently to verify that the marginals were perfectly reproduced. We also evaluate the conditional dependence of each attribute over other attributes, in effect deriving counts by category for each attribute. Finally, we measure the joint distributions on all possible combinations of data variables. This measure is applicable in low dimensional data but can be computationally intensive for high dimensional data.  In such instances, marginal and conditional joint distributions are recommended. We evaluate the success of the synthetic approach by the similarity score in probabilities of the distributions. We quantify the empirical distributions between the synthetic and true distributions with the Standard Root Mean Square Error (SRMSE), the fitness of the synthetic reconstruction using a measure of the Pearson Correlation Coefficient (corr) and the coefficient of determination (R$^2$). The standardized root mean squared error is defined by:

\begin{equation} \label{eq4}
SRMSE(\hat{\pi},\pi)=\frac{RMSE(\hat{\pi},\pi)}{\bar{\pi}}=\frac{\sqrt{\sum _i \cdots \sum _j(\hat{\pi} _i... _j -  \pi _i... _j)^2/N _b}}{\sum _i ... \sum _j \pi _{i...j}/N _b}
\end{equation}
where \textit{N$_b$} is the total number of agents; \textit{R$_{i..j}$} is the number of agents with attribute values i...j in the synthesized population, $\hat{\pi}$ and $\pi$ is the synthetic and true distribution respectively. 

\subsection{Pattern Analysis}
In this analysis, we adopt Principal component analysis (PCA) to measure the trends and patterns retained when synthesized data are reduced to fewer dimensions. The objective of PCA is to find the best summary of the data by reducing data using a geometric projection into a lower dimension. Using PCA, we can measure the variance of projected points and correlations between principal components. The output of the generative model should exhibit similarity in the variance and correlations between projected points. For each of the attributed sets in the synthesized data, numerical variables are normalized and categorical variables are converted to one-hot encoded representations. PCA is performed on all tuples of the generalized data. 

\section{Evaluation results}
In this section, we discuss the results achieved on performing the evaluation analysis. The generative model was developed and implemented with Python Keras with Tensorflow \cite{abadi2016tensorflow} backend support on a Windows 10 PC with Intel Core i7-2600 (8 Cores) and G-Force GTX 950.

\subsection{Statistical distribution comparison}
\begin{figure}%
    \centering
    
    \begin{subfigure}{0.5\textwidth}
    	\includegraphics[width=\linewidth, height=4.5cm]{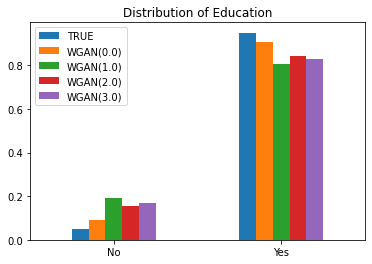} 
    	\caption{\textcolor{black}{GHG emission factor}}
    	\label{GHG EF}
    \end{subfigure}%
    \begin{subfigure}{0.5\textwidth}
	\includegraphics[width=\linewidth, height=4.5cm]{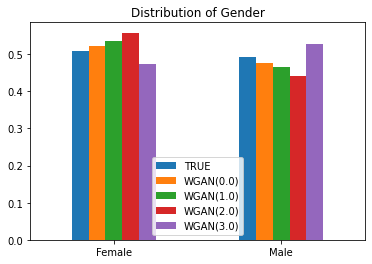} 
	\caption{\textcolor{black}{GHG emission factor}}
	\label{GHG EF}
\end{subfigure}%
    \caption{Comparison of marginals for attributes for True, WGAN and private WGAN representations}%
    \label{fig:marginals}%
\end{figure}

In this analysis, we compared the summary statistics on marginals, conditional and joint distributions for combinations of variables in the training and synthesized set. First, the marginals of the synthesized variables reproduced from the generative model produce the best approximation to the marginals of the true population. Figure \ref{fig:marginals} shows the marginals for 2 selected attributes from the true and synthesized population with varying privacy noise levels. It can be seen at WGAN(0.0) (no privacy noise added) that the simulated marginals sampler precisely reproduces the marginals of the training set, although a low error is observed due to the sampling bias persistent in the random selection of samples during training of the generative model. With an incremental addition of noise, the reproduced marginals of the simulated sampler are less precise compared to marginals of the training set and exhibit randomness in the error of prediction, which is not deterministic nor does it follow a monotonic pattern. As an example, in Figure \ref{fig:marginals}(a) while the prediction of ``Yes'' values is under-predicted, it can be seen at noise levels of 1.0, 2.0 and 3.0, that the under-prediction increases, but at noise level of 4.0, the under-prediction decreases. This is function of the level of randomness expected with the addition of noise such that an adversary cannot quantify if there will be a monotonic under-prediction or over-prediction. Marginals on the variable ``Gender'' shown in \ref{fig:marginals}(b) show similar characteristics.

\begin{figure}[tbh]
    \centering
    \includegraphics[width=0.65\textwidth]{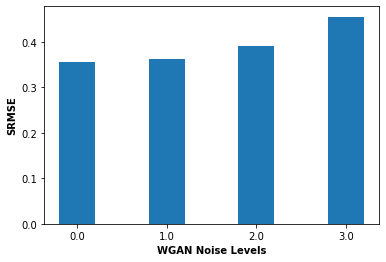}
    \caption{SRMSE on predictions for marginal distributions of synthetic agents using varying privacy noise levels}
    \label{fig:noiselevel}
\end{figure}

We study the goodness of fit by measuring the SRMSE of marginals observed between the true population and simulated populations at varying noise levels. In Figure \ref{fig:noiselevel}, we observe a monotonic pattern that depicting an increase in the SRMSE of predictions as privacy noise levels increase. An SRMSE of 0.356, 0.362, 0.390, 0.455 is observed at noise levels of 0.0, 1.0, 2.0 and 3.0 respectively. These results affirm the elastic nature of noise addition, a promise of differential privacy to control the difference between the distribution of the true and simulated by privacy budget.  

Similarly, we compare the fitting of empirical distributions on possible combinations of bivariate distributions on possible pairs of variables. The distributions of the conditional probabilities are computed as frequency tables where each bin corresponds to a specific combination of values between two data variables. We measure the SRMSE between true and simulation distributions for bivariate combinations including ``Permit vs Gender'', ``AgeGroup vs Gender'', ``AgeGroup vs Employed'' and ``Employed vs Gender''.  

\begin{table}[!h]
 \centering
\small
\begin{tabular}{|l|l|l|l|l|l|}
\hline
\textbf{NOISE LEVELS}  & \textbf{0.0} & \textbf{1.0} & \textbf{2.0} & \textbf{3.0} & \textbf{$\sigma_{mse}$} \\ \hline
p(Permit $\mid$ Gender)     & 0.432               & 0.576               & 0.464               & 0.497              & 0.053        \\ \hline
p(AgeGroup $\mid$ Gender)   & 0.614               & 0.703               & 0.527               & 0.588              & 0.063        \\ \hline
p(AgeGroup $\mid$ Employed) & 0.902               & 0.996               & 0.833               & 1.007              & 0.0714       \\ \hline
p(Employed $\mid$ Gender)   & 0.372               & 0.384               & 0.419               & 0.462              & 0.035        \\ \hline
\end{tabular}
\caption{SRMSE measured on bivariate conditional probabilities for synthetic agents using varying privacy noise levels. $\sigma_{mse}$ denotes the variance between SRMSE. } 
\label{tab:conditional-table}
\end{table}

As can be seen in Table \ref{tab:conditional-table}, introducing noise impacts prediction errors of the conditional probabilities of synthesized agents. Adding a noise of 1.0, 2.0 and 3.0, the SRMSE of p(Permit $\mid$Gender) increased from 0.432 to 0.576, 0.464 and 0.497 respectively. Similar to random perturbations in the marginals, the SRMSE does not exhibit a monotonic pattern hence suggesting randomness in the model prediction which makes it difficult for an adversary to estimate the pattern of prediction. For an example, at a noise of 1.0, SRMSE of p(AgeGroup $\mid$ Employed) increases to 0.996 but drops to 0.833 at noise of 2.0.

\begin{figure}
  \begin{subfigure}[t]{.4\textwidth}
    \centering
    \includegraphics[width=\linewidth]{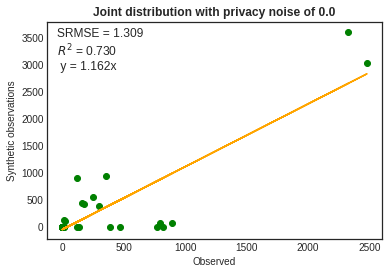}
    \caption{Full joint distribution of all variables at privacy noise level of \textbf{0.0}}
  \end{subfigure}
  \hfill
  \begin{subfigure}[t]{.4\textwidth}
    \centering
    \includegraphics[width=\linewidth]{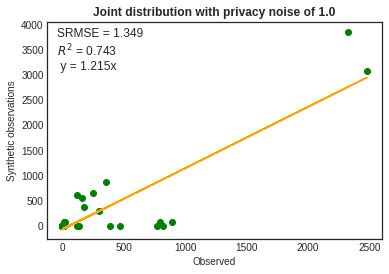}
    \caption{Full joint distribution of all variables at privacy noise level of \textbf{1.0}}
  \end{subfigure}

  \medskip

  \begin{subfigure}[t]{.4\textwidth}
    \centering
    \includegraphics[width=\linewidth]{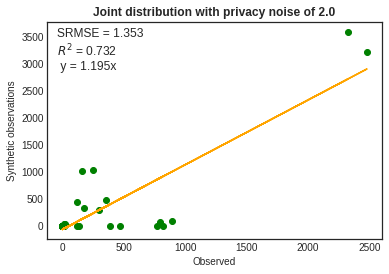}
    \caption{Full joint distribution of all variables at privacy noise level of \textbf{2.0}}
  \end{subfigure}
  \hfill
  \begin{subfigure}[t]{.4\textwidth}
    \centering
    \includegraphics[width=\linewidth]{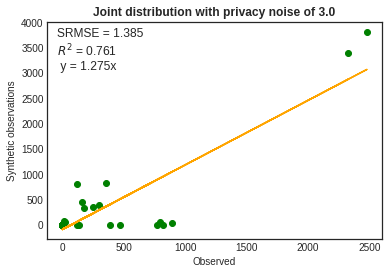}
    \caption{Full joint distribution of all variables at privacy noise level of \textbf{3.0}}
  \end{subfigure}
  \caption{Full joint distributions for all variables between observed and simulated counts.}
  \label{fig:joint}
\end{figure}

Finally, the joint probabilities were evaluated on all possible combinations of values on all data variables. We computed the frequency bins of all combinations for variables, p(AgeGroup, Employed, Gender, Educated). We show the fitting of the joint distributions for synthetic agents in Figure \ref{fig:joint}. The results of the generative model show a low performance in reproducing the joint probabilities of the synthetic agents. This inaccuracy of prediction can be attributed to the shared latent space that reduces the resolution of network parameters where multiple node branches having different dimensions are merged or compressed, as seen in the generator network of DP-CTGAN where the tabular and sequential branches are concatenated. This is evident in the joint distribution outputs in Figure \ref{fig:joint} predicting synthetic frequency counts of 0 where true population counts are about 800.  The line of fit exhibits a population balance observed between frequency bins of joint combinations for the synthetic agents. The mean square error of the prediction output increases with the magnitude of privacy noise added. For example, training at noise 1.0, SRMSE increases from 1.309 to 1.349. These marginal increases of SRMSE are consistent for larger noise additions as seen in Figure \ref{fig:joint} (c) and (d). The model exhibits the capability of maintaining a good joint distribution even with the introduction of noise.

\subsection{Tour length distributions}
\begin{figure}
  \begin{subfigure}[t]{.4\textwidth}
    \centering
    \includegraphics[width=\linewidth]{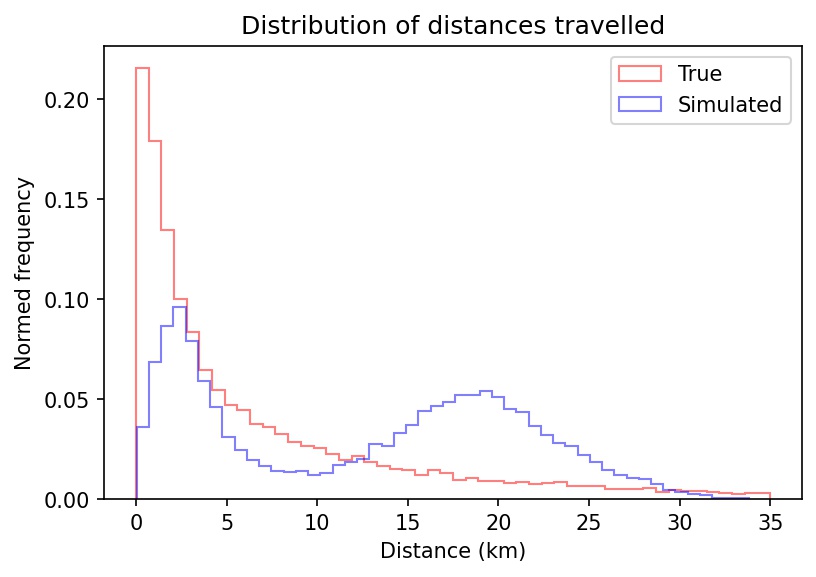}
    \caption{Marginal distribution of tour distances}
  \end{subfigure}
  \hfill
  \begin{subfigure}[t]{.4\textwidth}
    \centering
    \includegraphics[width=\linewidth]{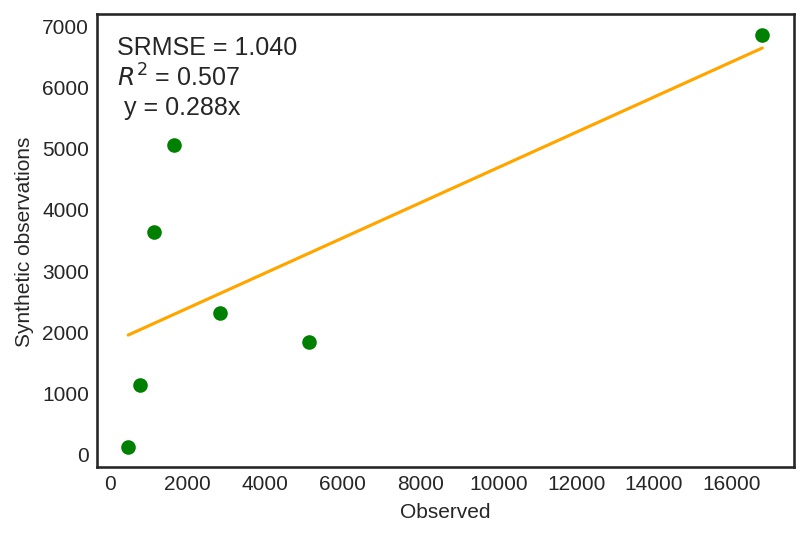}
    \caption{Line of fit on counts of tour distances}
  \end{subfigure}
  \caption{Comparison of distributions and fitting analysis between true and synthetic tour counts.}
  \label{fig:dist}
\end{figure}

In this analysis, we assess the similarity in sequence representations simulated by the generative model by calculating the distances between sequences of origin and destination geographic coordinates for the true and simulated sets. We assume an agent embarks on a tour composing a sequence of destinations based on an activity preference within time periods in a day. We evaluate the lengths of all tour segments made up of an origin and destination, and compare the joint distributions between the true and synthetic representations. In Figure \ref{fig:dist}a, we show the marginal distributions of computed tour distances for both the true and synthetic representations. It can be observed that the model under-predicts tour distances between 0km to 3km and 5km to 11km. Contrarily, the model over-predicts tour distances from 13km to 29km. While the memory capacity of the LSTM \cite{graves2012long} holds out the promise of learning correlations and representations for longer sequences, it can be observed that the model suffers the complexity of learning higher order correlations and long-range temporal dependencies needed for multiples features in longer timesteps. This drawback makes it difficult to learn longer sequences in complex generative architectures that could involve two or more networks learning with back-propagation. Oord et al. \cite{oord2016wavenet} have recently proposed a dilated convolution approach to address this drawback in longer generative sequences. To the best of our knowledge, this is the first attempt to implement such architecture in a multi-output with variable tour sequences thus this drawback needs further research to improve the prediction accuracy for long temporal dependencies. Similarly, the line of fit for tour length counts in Fig \ref{fig:dist}b shows an imbalance in the prediction, and recording a SRMSE of 1.040 and adjusted R-squared of 0.5. 

\subsection{Dimension reduction on principal components}
In this section, we explore analysis using principal component analysis (PCA) to identify the main axes of variance within the synthetic agents and evaluation on how the orthogonal variables correlate with the principal components. PCA constructs relevant features through linear combinations of the original variables. This construction is implemented by linearly transforming correlated features into lower dimensions of uncorrelated features using the eigenvectors of the correlation matrix. In this sense, the PCA undertakes an orthogonal transformation of the data into a reduced PCA space such that its derived components explain the most variance in the data.

\begin{figure}[!h]
  \begin{subfigure}[t]{.4\textwidth}
    \centering
    \includegraphics[width=\linewidth]{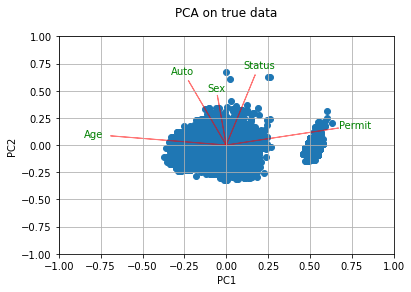}
    \caption{PCA on true population}
  \end{subfigure}
  \hfill
  \begin{subfigure}[t]{.4\textwidth}
    \centering
    \includegraphics[width=\linewidth]{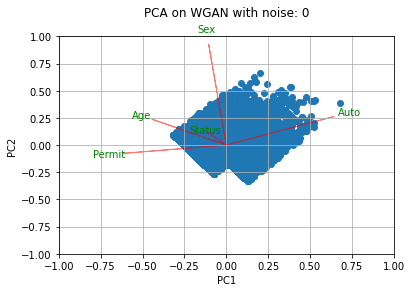}
    \caption{PCA on WGAN (0.0)}
  \end{subfigure}

  \medskip

  \begin{subfigure}[t]{.4\textwidth}
    \centering
    \includegraphics[width=\linewidth]{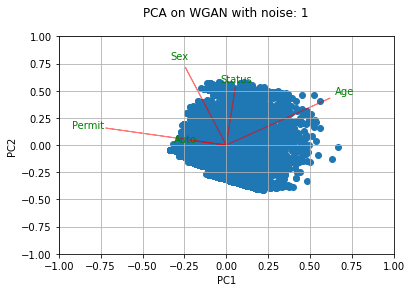}
    \caption{PCA on WGAN (1.0)}
  \end{subfigure}
  \hfill
  \begin{subfigure}[t]{.4\textwidth}
    \centering
    \includegraphics[width=\linewidth]{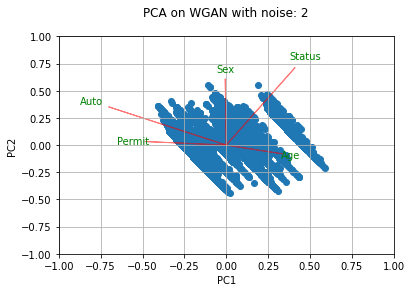}
    \caption{PCA on WGAN (2.0)}
  \end{subfigure}
  \caption{Principal component analysis on true and synthesized agents with varying privacy noise levels.}
  \label{fig:pca}
\end{figure}

In Figure \ref{fig:pca}, the ``biplots" and ``loadings" plot show the association between the orthogonal variables and their clusters. Variables \textit{age, sex and status} are highly correlated and form a cluster in representations of the true set synthesized with noise of 0. With the introduction of noise with magnitudes 1 and 2, the angles between these 3 variable vectors widen suggesting lower associations between them. Similarly, Principal component 2 (PC2) shows a strong correlation with variables \textit{age, sex and status} in Figure \ref{fig:pca}(a) and (b), and exhibit  high positive loadings, suggesting PC2 will increase when the scores of the three variables increase. As noise is introduced, and as can be seen in Figures \ref{fig:pca}(c) and (d), these correlations decay.  

In summary, the PCA analysis has shows that the introduction of noise distorts associations and correlations in representations of synthetic agents. This shows, that the magnitude of noise controls the level of distortions that influence the correlations in synthesized representations. 

\subsection{Adversarial predictions on target models with knowledge on parameters}
\begin{figure}
  \begin{subfigure}[t]{.4\textwidth}
    \centering
    \includegraphics[width=\linewidth]{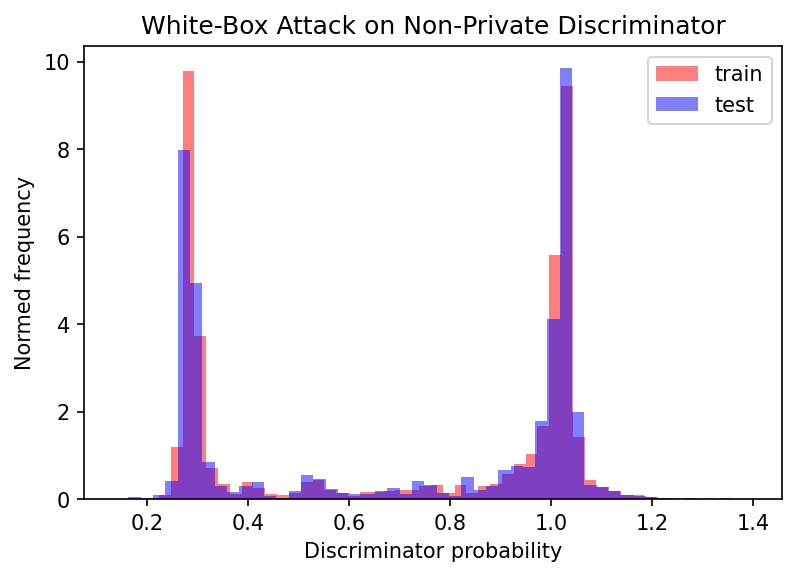}
    \caption{Attack with no private training}
  \end{subfigure}
  \hfill
  \begin{subfigure}[t]{.4\textwidth}
    \centering
    \includegraphics[width=\linewidth]{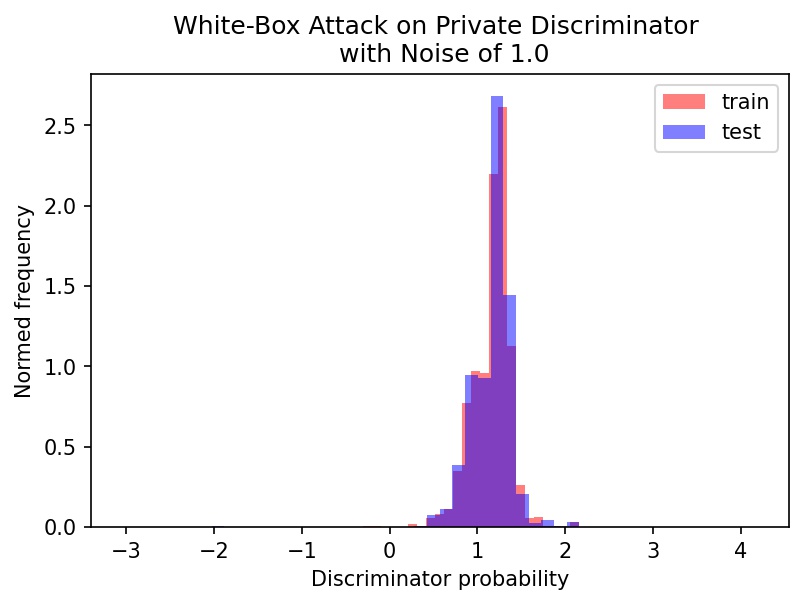}
    \caption{Attack with private training at noise level of \textbf{1.0}}
  \end{subfigure}

  \medskip

  \begin{subfigure}[t]{.4\textwidth}
    \centering
    \includegraphics[width=\linewidth]{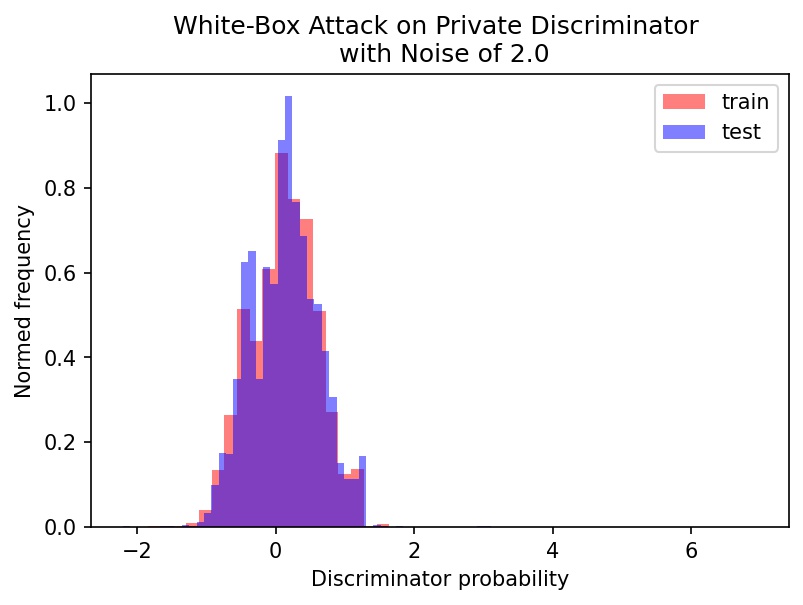}
    \caption{Attack with private training at noise level of \textbf{2.0}}
  \end{subfigure}
  \hfill
  \begin{subfigure}[t]{.4\textwidth}
    \centering
    \includegraphics[width=\linewidth]{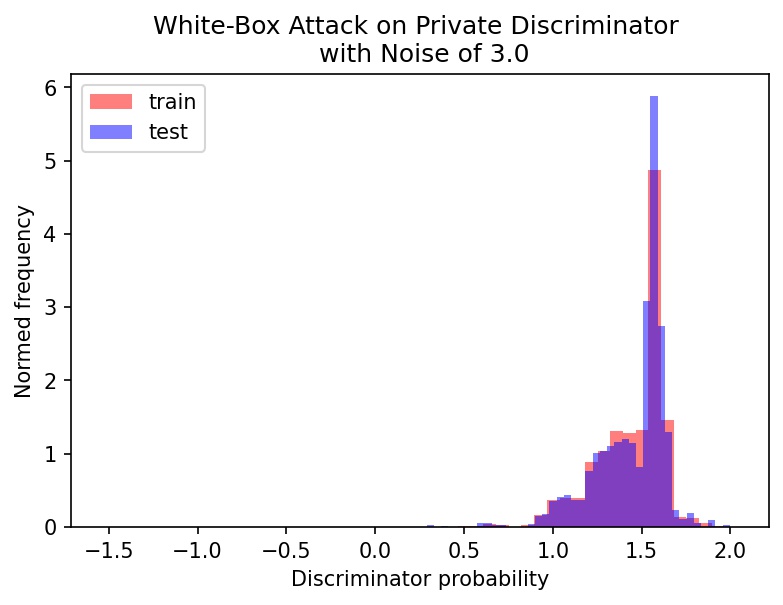}
    \caption{Attack with private training at noise level of \textbf{3.0}}
  \end{subfigure}
  \caption{White-box attacks on trained discriminator model with varying noise levels.}
  \label{fig:attack}
\end{figure}

In the previous sections we've demonstrated how noise can be incorporated into agent simulation with the DP-CTGAN. The noise introduced, however, does not in itself provide information on the degree to which this addition can protect the privacy of individuals in the original sample data. In order to evaluate this, we need to consider the degree of privacy protection that simulated DP-CTGAN data can provide in the context of deliberate adversarial attacks.
To do so, we assume that an adversary has knowledge on the model parameter of the target model, \textit{the trained discriminator}. Fundamentally, the objective of the discriminator is to distinguish between true or fake samples. This means samples in the training set of the model should result in higher predictive scores than validation sets. As observed in Fiure \ref{fig:attack}a, a bimodal distribution with two peaks having means approximating to zero (0) and one (1) are recorded for prediction scores from the non-private trained discriminator model. The bimodal distribution affirms the accuracy of classification outputs from the target model. This implies that when an adversary has data on the entire population including the training sets at his disposal, he can predict with high confidence level whether a sample data point was used in the training of the model. It can also be seen that the discriminator does not perfectly classify but shows traces of proportions of validation sets also having a high score and vice versa. This occurs because of the similarity in features for both the training and validation sets thus members of the sample population have near similar attributes. In Figure \ref{fig:attack}b, a unimodal distribution centered around 1 is derived for both the training and validation sets. The two peaks of the bimodal distribution as expected for the classification by a discriminator diminishes into a unimodal. This means the target model fails to classify between the samples that were used for training and validation. In this sense, the adversary cannot exploit the target model to infer if a data point was used in the training. This is the promise of differential private training by stochastic gradient descent \cite{abadi2016deep}. We perform sensitivity tests on differing noise levels to privately train the target model as shown in Figures \ref{fig:attack}c and d. The results show unimodal prediction score in both cases, suggesting the failure of the adversary to correctly classify training or validation sets in either case. In summary, privacy protection is guaranteed on the differential-private trained models against any attacks in the case that an adversary has access to the parameters of the target model. That is, any record of a person that participated in the training of the model cannot be confidently identified by an adversary who has access to the model parameters.

\section{Discussions and Conclusions}
We develop and demonstrate the use of the novel Differentially-Private Composite Travel Generation Adversarial Network (DP-CTGAN) for full activity diary synthesis, accounting for both accuracy of the population synthesized and the privacy of the individuals in the estimation datasets. The population synthesized using DP-CTGAN can be directly used as the base-input for the modern activity-based travel demand modelling systems. Moreover, the DP-CTGAN shows success in simulating activity diaries composed of multiple outputs including structured socio-economic features and sequential trip activities in a differentially private manner. To implement privacy-by-design, the generative model was trained by bounded gradients of data points with added Gaussian noise propagated from the discriminator. The outputs of the synthetic agents shown appreciable similarity in statistical properties to the true population such that lower SRMSE is observed on the joint distributions for the prediction of the  simulated agents.

We evaluate the similarity in statistical properties by comparing the marginals, conditionals and joint probabilities of synthetic representations of varying privacy noise levels to the true distribution. The results show a consistent output that provides a level of randomness influenced by the addition of noise such that the query probabilities differed by the specified level of noise. Observed root mean square errors increased with the addition of more noise on a line of fit of synthetic probabilities to the true distribution.

To the best of our knowledge, this is the first work that adopts a deep learning approach to simulate population synthesis of trip data having multiple outputs with different dimensions (i.e., structured and sequential) and guarantees privacy protection. While the model promises the reproducibility of sequential activities, our approach suffers limitations on training multiple features in long sequence generation using LSTM. The LSTM approach was not effective in producing accurate samples for higher dimensional and longer sequences. In practice, most existing literature has implemented similar sequence generations with one dimensional data. We foresee this drawback as a research interest that needs to be further studied to develop models that are robust and efficient in generation of multiple features and lengthy sequences as required by travel trajectories. Also, the approach of training deep learning with differential privacy as implemented in this paper makes it intractable to sample geographic location points whose positional accuracies are of high priority in transportation modelling. During training with differential privacy, the magnitude of random noise injected could perturb the normalized location coordinates making them less useful for microsimulation. In our future research, we will perform detailed sensitivity analysis on hyper-parameters that can control the injection of noise on accuracy and precision demanding variables like positional coordinates when used in synthesis approaches.

Further research will be done to improve on the fitting for true population especially in the context of the conditional and full joint probabilities of the empirical distributions. Also, we will perform state-of-art adversarial attacks \cite{chen2019gan, hayes2019logan} on the generative model to test on its robustness against attacks such as when an adversary with enough background knowledge seeks to infer whether information of an individual was used in the training data or when the adversary seeks to learn something new about an individual from the synthesized outputs. 

Finally, we will extend this research into designing model frameworks that support privacy-by-design techniques in generating usable location and activity diary sequences for microsimulation. While little research is available on benchmarks for evaluating multidimensional sequences, we will further this work into defining and developing such metrics in the context of location-aware privacy protection.

\bibliographystyle{unsrt}  
\bibliography{references}

\end{document}